\pdfoutput=1
\documentclass[11pt]{article}

\usepackage[]{acl}

\usepackage{times}
\usepackage{latexsym}

\usepackage[T1]{fontenc}
\usepackage[utf8]{inputenc}

\usepackage{microtype}

\usepackage{graphicx}
\usepackage{arydshln}
\usepackage{xcolor}

\title{Autoregressive Score Generation for Multi-trait Essay Scoring}

\author{Heejin Do$^{1}$ \ \ \ Yunsu Kim$^{2}$ \ \ \ Gary Geunbae Lee$^{1, 3}$ \\
{\normalsize $^{1}$Graduate School of AI, POSTECH, Republic of Korea}\\
{\normalsize $^{2}$aiXplain Inc. Los Gatos, CA, USA}\\
{\normalsize $^{3}$Department of CSE, POSTECH, Republic of Korea}\\
\texttt{\small \{heejindo, gblee\}@postech.ac.kr \ yunsu.kim@aixplain.com}
}

\begin{document}
\maketitle
\begin{abstract}
%Automated essay scoring (AES), which assigns numerical scores to essays, has been chiefly approached in classification or regression ways.

% Considering the increasing demands of predicting multiple trait scores for diverse feedback yet still lacking research, we aim to enhance multi-trait AES. Specifically, we propose an autoregressive generation process that sequentially predicts the text-transformed multi-trait scores by leveraging the pre-trained language model, T5.

% While demand for multi-trait scoring has increased, research in this area remains limited. Thus, we propose ArTS, an autoregressive prediction model specifically designed for multi-trait scoring. 
Recently, encoder-only pre-trained models such as BERT have been successfully applied in automated essay scoring (AES) to predict a single overall score. However, studies have yet to explore these models in multi-trait AES, possibly due to the inefficiency of replicating BERT-based models for each trait. Breaking away from the existing sole use of \textit{encoder}, we propose an autoregressive prediction of multi-trait scores (ArTS), incorporating a \textit{decoding} process by leveraging the pre-trained T5. Unlike prior regression or classification methods, we redefine AES as a score-generation task, allowing a single model to predict multiple scores. During decoding, the subsequent trait prediction can benefit by conditioning on the preceding trait scores. Experimental results proved the efficacy of ArTS, showing over 5\% average improvements in both prompts and traits.
%Narrowing the gap between single-holistic scoring and multi-trait scoring

%Automated essay scoring (AES) has recently been tackled with pre-trained models, such as BERT. However, solely encoder-only models are applied to output a score in a classification or regression manner. This paper redefines AES as a generation task that autoregressively decodes the scores. Considering growing demands for multi-trait scoring to provide diverse feedback yet still lack research, we propose an autoregressive prediction, especially for the multi-trait scoring of essays (ArTS). We sequentially predict the text-transformed multi-trait scores by leveraging the pre-trained language model, T5. In the decoding process, the subsequent trait-scoring tasks can be assisted by attending to the precedent trait scores. Experimental results prove the efficacy of ArTS, showing average 5.2\% and 5.7\% improvements on prompt- and trait-wise, respectively.

%Experimental results prove that our method significantly improves the multi-trait scoring performance.
\end{abstract}

\section{Introduction}
% In this paper, we redefine AES as a generation task that autoregressively decodes the scores, introducing a novel text-to-text AES.

%, for intensive essay representation to predict a numeric score,

Automated essay scoring (AES) is a prominent task to efficiently assess large volumes of essays. Currently, there is a growing trend in holistic AES to use pre-trained BERT-based models, showing promising results \cite{rodriguez2019language, mayfield2020should, beseiso2020empirical, yang2020enhancing, wang2022use}. However, these models have yet to be explored in multi-trait AES, which evaluates essays on diverse rubrics, possibly due to the inefficiency of duplicating encoders for different traits.

Existing multi-trait scoring approaches \cite{mathias2020can, ridley2021automated, kumar2022many, do2023prompt} typically adopted holistic scoring models \cite{taghipour2016neural, dong2017attention}, adding multiple linear layers or separate trait-specific layers for different traits. However, achieving multi-trait AES as a holistic method overlooks the trait dependencies, and constructing separate trait-specific modules is resource-inefficient, leading to inferior qualities in data-scarce traits. These limitations highlight the need for optimized multi-trait strategies.

In this paper, we propose autoregressive multi-trait scoring of essays (ArTS), which incorporates the decoding process by leveraging a pre-trained language model, T5 \cite{2020t5}. Moving beyond the conventional sole reliance on the encoder, we introduce a novel text-to-text AES framework. Unlike existing regression or classification approaches to output a separate numeric value, we aim at precise sequence generation by considering multi-trait scores as an entire sequence; thus, a single model can yield multi-score predictions. ArTS employs causal self-attention to capture the intrinsic relations of the traits by sequentially predicting text-transformed trait scores. The autoregressive generation allows the subsequent trait prediction to benefit from referencing preceding trait scores.

ArTS remarkably outperformed the baseline model on the ASAP and ASAP++ \citep{mathias2018asap++} datasets. Ablation studies and additional discussions of trait order further verify our method. Furthermore, ArTS achieved training efficiency by using a single model to generate multiple predictions across all prompts, avoiding the duplication of the same modules. Codes and datasets are available on Github\footnote{\url{https://github.com/doheejin/ArTS}}.
% 여기서 각각 어떻게 도움되는지 result 쓴 다음에 구체화하기 

\begin{figure*}[t]
    \centering
    \includegraphics[width=16cm]{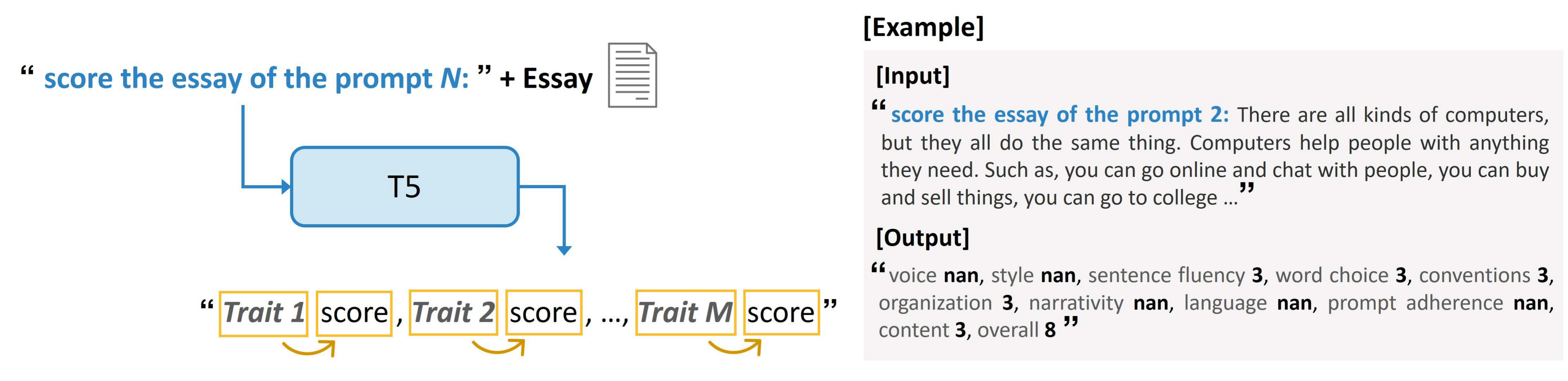}
    \caption{Proposed autoregressive multi-trait essay scoring by the fine-tuning of the T5. The example is an essay written for prompt 1, which has labeled scores for six traits. Unlabeled trait scores in the prompt are set as \textit{nan}.}
    \label{fig1}
\end{figure*}

\section{Related Work}

Early studies of AES mainly focused on holistic essay scoring that only predicts the overall score and already achieved high assessment performance \cite{dong2016automatic, taghipour2016neural, dong2017attention, uto2020neural, wang2022use}. In contrast, multi-trait scoring has been studied for detailed assessments lately, yet showing far-lagged quality. Holistic scoring structures are typically employed either for a trait-shared model followed by multiple linear layers \cite{hussein2020trait} or for multiple trait-specific layers \cite{mathias2020can, ridley2021automated, kumar2022many, he2022automated,  do2023prompt}. In particular, \citet{kumar2022many} designed auxiliary trait-specific layers to assist primary trait scoring, achieving competitive results. However, to predict $m$ trait scores, $m$ different models containing $m$ duplicated trait-specific layers are required, which is resource-inefficient. Moreover, the notable quality gap between trait scoring and holistic scoring highlights the need for advanced multi-trait AES.
% surpassing the extension of holistic scoring perspectives

% AES에 pre-trained model 쓴 사례
%%% 우리는 최초로 seq-2-seq을 제안한다. 

% 기존의 trait scoring 방법론들
%%% 우리는 M개의 다른 모델 안 쓰고 단 하나의 모델로도 좋은 성능 낼 수 있기 때문에 자원 효율성과 성능 효과성 두 마리 토끼를 잡았다. 

%Transformer-based pre-trained models like BERT \cite{devlin2018bert} and GPT \cite{brown2020language} excel across various tasks by capturing rich semantic and syntactic information from large-scale corpora. 

Transformer-based pre-trained models such as BERT \cite{devlin2018bert} and GPT \cite{brown2020language} excel across various tasks by capturing rich semantic and syntactic information via training on large-scale corpora. Recently, some studies have applied them to \textit{holistic} AES \cite{rodriguez2019language, mayfield2020should, beseiso2020empirical, yang2020enhancing, wang2022use}, contributing to a notable leap in the \textit{holistic} scoring. However, they only employ encoder-only models to predict a numeric value without considering the decoder. Moreover, those BERT-based models have not been extended to multi-trait scoring, possibly due to the efficiency concerns (e.g., predicting an \textit{Overall} score with a BERT-based model of 110M parameters took 113 hours \citep{kumar2022many}; accordingly, predicting \textit{m} traits would require \textit{m} times the parameters and the time). In contrast, we leverage the potential capacities of autoregressive decoding to efficiently score multiple traits with a single model, suggesting a new perspective to address AES as a text generation task instead of a classification or regression.

%Unlike previous approaches, we utilize autoregressive decoding to efficiently score multiple traits with a single model, proposing a novel approach to view AES as a text generation task rather than a classification or regression.

%This introduces a new perspective to approach AES not as a classification or regression task for numeric prediction but as a text-to-text generation task.

%This novel approach shifts the perspective from viewing AES as a classification or regression task for numeric prediction to a text-to-text generation task
%, opening new avenues for advanced multi-trait scoring.

% 기존의 encoder 기반 접근과 다르게, 우리는 leveraging the potential capacities of autoregressive decoding 함으로써 multi-trait scoring에서도 pre-trained 모델의 효율적인 도입을 가능하게 한다. 이로써 AES를 numeric prediction을 위한 regression task로 보는 기존의 접근에서 벗어나 text-to-text generation task로 접근하는 새로운 시각을 제시한다.  

%Unlike prior works, we propose a novel paradigm for AES by adopting a text-to-text framework, leveraging the potential capacities of autoregressive decoding.

% 더군다나 multi-trait scoring에는 bert기반 model이 쓰인 적이 없고 이는 여러 duplicate bert 모델을 사용하는 것이 ~~ 
% 이 부분을 뒤에서 서술하기 

\section{Autoregressive Essay Multi-trait Scoring (ArTS)}

To predict multiple trait scores in an auto-regressive manner, we fine-tune the pre-trained encoder-decoder language model, T5. Specifically, we treat AES as a generation task to predict a single sequential text rather than multiple numeric values for traits. Subsequently, we extract each trait score from the generated text comprising the predicted trait scores along with trait names (Figure~\ref{fig1}).

\subsection{Fine-tuning T5}
T5 has achieved competitive performance in numerous natural-language processing tasks by handling various tasks using a text-to-text approach. One of the trained tasks of T5 is semantic textual similarity (STS), which is a regression task predicting a float-type similarity value between two texts. Given that T5 has been pre-trained to output a text-formed numeric value for the STS, we assume that fine-tuning the model to output an essay score will yield precise prediction. Instead of individually predicting trait scores with multiple models, our goal is to generate all trait scores with a single autoregressive prediction, thus achieving both time and resource efficiency. Using one integrated model can avoid unnecessary duplication of the same distinct models.

\begin{table}[t]
\centering
\scalebox{
0.7}{
\begin{tabular}{c|c|l}
\hline
Prompt & \# Essays & Traits\\
\hline
1 & 1785 & Over, Content, WC, Org, SF, Conv \\
2 & 1800 & Over, Content, WC, Org, SF, Conv \\
3 & 1726 & Over, Content, PA, Nar, Lang \\
4 & 1772 & Over, Content, PA, Nar, Lang \\
5 & 1805 & Over, Content, PA, Nar, Lang \\
6 & 1800 & Over, Content, PA, Nar, Lang \\
7 & 1569 & Over, Content, Org, Conv, Style \\
8 & 723 & Over, Content, WC, Org, SF, Conv, Voice \\
\hline
\end{tabular}}
\caption{\label{tab1}
Composition of the ASAP/ASAP++ combined dataset. The prompt is an instruction that defines the writing theme. Over: \textit{Overall}, WC: \textit{Word Choice}, Org: \textit{Organization}, SF: \textit{Sentence Fluency}, Conv: \textit{Conventions}, PA: \textit{Prompt Adherence}, Nar: \textit{Narrativity}, Lang: \textit{Language}.}
\end{table}

\begin{table*}[t]
\centering
\scalebox{
0.73}{
\begin{tabular}{l|ccccccccccc|c}
\hline
& \multicolumn{11}{c|}{\textbf{Traits (←)}} & \\
\hline
\textbf{Model} &  Overall & Content & PA & Lang & Nar & Org & Conv & WC & SF & Style & Voice & AVG↑ (SD↓)\\
\hline
HISK & 0.718 & 0.679 & 0.697 & 0.605 & 0.659 & 0.610 & 0.527 & 0.579 & 0.553 & 0.609 & 0.489 & 0.611 (-) \\
STL{\small-LSTM} & 0.750 & 0.707 & 0.731 & 0.640 & 0.699 & 0.649 & 0.605 & 0.621 & 0.612 & 0.659 & 0.544 & 0.656 (-) \\
MTL{\small-BiLSTM} & \textbf{0.764} & 0.685 & 0.701 & 0.604 & 0.668 & 0.615 & 0.560 & 0.615 & 0.598 & 0.632 & \textbf{0.582} & 0.638 (-) \\
\hline
\textbf{ArTS} (Ours) & 0.754 & \textbf{0.730} & \textbf{0.751} & \textbf{0.698} & \textbf{0.725} & \textbf{0.672} & \textbf{0.668} & \textbf{0.679} & \textbf{0.678} & \textbf{0.721} & 0.570 & \textbf{0.695} (±0.018) \\
\textcolor{gray}{\textbf{ArTS}{\small-w/o \textit{Pr}}} & \textcolor{gray}{0.690} & \textcolor{gray}{0.723} & \textcolor{gray}{0.751} & \textcolor{gray}{0.691} & \textcolor{gray}{0.725} & \textcolor{gray}{0.655} & \textcolor{gray}{0.656} & \textcolor{gray}{0.644} & \textcolor{gray}{0.648} & \textcolor{gray}{0.673} & \textcolor{gray}{0.530} & \textcolor{gray}{0.671 (±0.033)} \\
\hline
\end{tabular}}
\caption{\label{tab2}
Average QWK scores across all prompts for each \textbf{trait}. The left arrow (←) indicates the direction of the trait prediction. \textit{SD} is the five-fold averaged standard deviation. ArTS{\small-w/o \textit{Pr}} (shown in gray) represents the ablation results without the prompt indication. Further, \textbf{bold} text denotes the highest value, excluding ablation results.}

\smallskip
\smallskip

\scalebox{
0.73}{
\begin{tabular}{l|cccccccc|c}
\hline
& \multicolumn{8}{c|}{\textbf{Prompts}} & \\
\hline
\textbf{Model} & 1 & 2 & 3 & 4 & 5 & 6 & 7 & 8 & AVG↑ (SD↓) \\
\hline
HISK & 0.674 & 0.586 & 0.651 & 0.681 & 0.693 & 0.709 & 0.641 & 0.516 & 0.644 (-) \\
STL{\small-LSTM} & 0.690 & 0.622 & 0.663 & 0.729 & 0.719 & 0.753 & 0.704 & 0.592 & 0.684 (-) \\
MTL{\small-BiLSTM}  & 0.670 & 0.611 & 0.647 & 0.708 & 0.704 & 0.712 & 0.684 & 0.581 & 0.665 (-) \\
\hline
\textbf{ArTS} (Ours) & \textbf{0.708} & \textbf{0.706} & \textbf{0.704} & \textbf{0.767} & \textbf{0.723} & \textbf{0.776} & \textbf{0.749} & \textbf{0.603}& \textbf{0.717} (±0.025) \\
\textcolor{gray}{\textbf{ArTS}{\small-w/o \textit{Pr}}} & \textcolor{gray}{0.709} & \textcolor{gray}{0.645} & \textcolor{gray}{0.703} & \textcolor{gray}{0.769} & \textcolor{gray}{0.679} & \textcolor{gray}{0.769} & \textcolor{gray}{0.722} & \textcolor{gray}{0.566} & \textcolor{gray}{0.695 (±0.036)} \\
\hline
\end{tabular}}
\caption{\label{tab3}
Average QWK scores across all traits for each \textbf{prompt}.}
\end{table*}

Particularly, we add the prefix \textit{"score the essay of the prompt N:"} in front of each essay as the input and concatenate trait name and trait score sets sequentially from the least to the most data labels with a comma (,) separation (Figure \ref{fig1}). We hypothesize that providing the prompt number, \textit{N}, allows more accurate guidance. Note that traits not labeled in the corresponding prompt are trained to predict \textit{nan} values. Including \textit{nan} values might allow the model to generate a consistent output form regardless of the prompt, leading to more reliable predictions. In particular, the model predicts traits in the following order: \textit{Voice}, \textit{Style}, \textit{SF}, \textit{WC}, \textit{Conv}, \textit{Org}, \textit{Nar}, \textit{Lang}, \textit{PA}, \textit{Content}, and \textit{Overall} (Table~\ref{tab1}). By predicting peripheral trait scores first, which are assessed in fewer prompts, and more comprehensive trait scores later, which are rated in more prompts, we reflect the actual scoring process. For example, the \textit{Overall} score is labeled in all prompts and highly influenced by other traits, whereas the \textit{Voice} score is only evaluated in prompt 8 (Table~\ref{tab1}) and is relatively independent of other traits. The causal self-attention of the transformer decoder enables subsequent trait-scoring tasks to attend to prior predicted trait scores; thus, the later order of dependent and general traits is natural.

\subsection{Score extraction}
With the fine-tuned model, we predict and generate a text for each essay containing predicted multiple trait scores along with the trait names. Then, we extract all trait scores keyed by their name. Multiple trait scores are obtained with a single model at one inference time, eliminating the inconvenience of multiple-model training and inference. For accurate measurement, we exclude all predictions of traits whose ground truth is a \textit{nan} value.

\section{Experiment} \label{sec4}
% 어떤 실험들 했는지 쓰기

\paragraph{Datasets and settings}
For the main experiment, we employ the widely used ASAP\footnote{https://www.kaggle.com/c/asap-aes} and ASAP++\footnote{https://lwsam.github.io/ASAP++/lrec2018.html} \citep{mathias2018asap++} datasets comprising English essay sets for eight prompts written by American 7--10-grade high-school students. The \textit{Overall} score is available for all essays in the ASAP dataset; however, trait scores are only labeled for essays of prompts 7 and 8. Therefore, the ASAP++ dataset providing rated trait scores for all prompts is jointly used (Table~\ref{tab1}).
%All essays have a labeled overall score, but the trait scores are only labeled for essays of prompts 7 and 8. Thus, the ASAP++ dataset providing rated trait scores for all prompts' essays is also used (Table~\ref{tab1}). 
In addition, we experiment on the Feedback Prize\footnote{https://www.kaggle.com/competitions/feedback-prize-english-language-learning} data of argumentative essays written by American 6--12-grade students. It has six labeled trait scores without prompt division: \textit{Cohesion}, \textit{Syntax}, \textit{Vocabulary}, \textit{Phraseology}, \textit{Grammar}, and \textit{Conventions}.
% 우리 모델이 더 범용적으로 적용됨을 증명하기 위해 우리는 
% ASAP 데이터는 8개의 prompt에 대해 작성된 에세이들셋들로 구성되어 있고, 에세이들은 prompt별로 다른 trait들에 대한 점수가 매겨진다. 

We utilize the T5-Base \cite{2020t5} model, which is pre-trained on the Colossal Clean Crawled Corpus. For fine-tuning, we employ Seq2SeqTrainer by setting evaluation steps as 5000, early stopping patience as 2, batch size as 4, and total epoch as 15. \texttt{A100-SMX4-8} GPU is used.
% 우리는 

\begin{table*}[t]
\centering
\scalebox{
0.73}{
\begin{tabular}{l|ccccccccccc|c}
\hline
& \multicolumn{11}{c|}{\textbf{Traits}} & \\
\hline
\textbf{Model} & Overall & Content & PA & Lang & Nar & Org & Conv & WC & SF & Style & Voice & AVG↑ (SD↓)\\
\hline
\textbf{ArTS} (←) & \textbf{0.754} & \textbf{0.730} & {0.751} & \textbf{0.698} & \textbf{0.725} & \textbf{0.672} & \textbf{0.668} & \textbf{0.679} & \textbf{0.678} & {0.721} & \textbf{0.570} & \textbf{0.695} (±0.018)\\
\hline
ArTS\textit{-rev} (→) & 0.739 & 0.724 & 0.749 & 0.687 & 0.718 & 0.667 & 0.658 & 0.660 & 0.666 & 0.711 & 0.562 & 0.686 (±0.021) \\
ArTS\textit{-ind} & 0.723 & 0.717 & \textbf{0.752} & 0.695 & 0.713 & 0.649 & 0.659 & 0.662 & 0.675 & \textbf{0.722} & 0.548 & 0.683 (±0.053) \\
\hline
 \end{tabular}}
\caption{\label{tab4}
Comparison results averaged by traits. ArTS\textit{-rev} (→) predicts traits in reverse order, and 11 different ArTS\textit{-ind} models predict each trait individually. The left (←) and right (→) arrows denote the direction of prediction.}
\end{table*}

\begin{table}[t]
\centering
\scalebox{
0.72}{
\begin{tabular}{c|c|c|c|c|c|c|c}
\hline
& \multicolumn{6}{c|}{\textbf{Traits (→)}} & \\
\hline
\textbf{Model}  & Conv & Gram & Phr & Voc & Syn & Coh & AVG \\
\hline
MTL* &  0.527 & 0.484 & 0.505 & 0.519 & 0.507 & 0.462 & 0.501 \\
\textbf{ArTS} & \textbf{0.659} & \textbf{0.659} & \textbf{0.639} & \textbf{0.594} & \textbf{0.628} & \textbf{0.590} & \textbf{0.628} \\
\hline
\end{tabular}}
\caption{\label{tab5}
Experiments with the Feedback Prize dataset. Each value is the five-fold average QWK score (Conv: \textit{Conventions}, Gram: \textit{Grammar}, Phr: \textit{Phraseology}, Voc: \textit{Vocabulary}, Syn: \textit{Syntax}, Coh: \textit{Cohesion}).}
\end{table}

\paragraph{Evaluation and validation}
For evaluation, we use the quadratic weighted kappa (QWK) \cite{cohen1968weighted}, the official metric of the dataset. QWK is well-known for effectively capturing the distance between human-rated and model-predicted scores. We use five-fold cross-validation with the same split as that of \citet{taghipour2016neural}, as in the baseline multi-task learning (MTL) \cite{kumar2022many}, reporting five-fold averaged results. We short-list two models based on the validation loss and select the final model with the best validation result. As suggested by \citet{taghipour2016neural}, we calculate QWK separately for each prompt to avoid excessively high scores when using the whole set (e.g., $0.99$ QWK for \textit{Overall} with ArTS), providing both prompt- and trait-wise averaged results.

%As pointed out in \citet{taghipour2016neural}, calculating QWK for the whole test set instead of calculating each prompt separately leads to extremely high (e.g., 0.99 with our method) QWK scores. Thus, we measure the QWK of multiple traits separately for each prompt, reporting both prompt- and trait-wise averaged results.

\section{Results}\label{sec5}
Our model is primarily compared with the baseline MTL-{\small BiLSTM} model \cite{kumar2022many}, multi-task learning where auxiliary multi-trait scoring tasks aid holistic scoring (Table~\ref{tab2}). In addition, we compare our model to the HISK and STL-{\small LSTM} models, which were mainly compared to MTL. HISK is a histogram intersection string kernel with a support vector regressor \cite{cozma-etal-2018-automated}, and STL-{\small LSTM} is LSTM-CNN-based model \cite{dong2017attention}; both models are individually applied for each trait scoring. Trait-scoring results are only presented with a graph \cite{kumar2022many}; thus, we contacted the authors and obtained exact values.

\paragraph{Main results} 
ArTS exhibits a significantly improved performance, showing over 5\% average improvements in both prompt- and trait-wise results (Table~\ref{tab2},~\ref{tab3}). %The \textit{Overall} scoring is slightly decreased, which might be attributed to our model's general focus on all traits, as opposed to baseline models designed primarily for overall scoring.
A slight decrease in \textit{Overall} trait could be attributed to our model's general focus on all traits, as opposed to baseline models designed primarily for overall scoring. For syntactic traits (\textit{Org}, \textit{Conv}, \textit{WC}, \textit{SF}), which evaluate the structure or grammatical aspects of essays, the performance increases by an absolute 5.7--10\%. This highlights that leveraging ArTS facilitates capturing essays' syntactic aspects, even with few datasets. Notably, the \textit{Conv} trait, the most inferior trait on the baseline, shows the greatest improvement with ArTS. Remarkably enhanced semantic traits (\textit{Content}, \textit{PA}, \textit{Lang}, \textit{Nar}) further imply that our autoregressive approach adeptly encapsulates the contextual facets of writing. Further, \textit{Style} and \textit{Voice} traits with severely lacking (1569, 723) samples show approximately 9\% advancement and a slight reduction, respectively, implying the overcoming of low-resource settings.

\paragraph{Prompt number guidance} 
We conducted an ablation study to investigate the effect of providing a prompt number in training. ArTS{\small-w/o \textit{Pr}} (Table~\ref{tab2},~\ref{tab3}) is the model results fine-tuned with the prefix \textit{"score the essay:"} without the prompt number. The results indicate that clearly guiding the model with the essay's prompt number noticeably assists the scoring.

% ASAP 데이터를 사용한 모델보다 더 높은 폭의 성능 향상은 trait 구성이 동일한 셋팅에서는 우리 모델이 더 효과가 있음을 암시해줍니다.   

% To investigate 
\paragraph{Trait prediction order}
To investigate the effect of the trait prediction sequence, we fine-tune T5 with the reverse order (ArTS-\textit{rev}). Improved results when predicting general traits later in the sequence than the reverse reflect the real-world scoring, where comprehensive trait scores often rely on the other traits \cite{lee2010toward}. In addition, we compare ArTS with the individual trait models (Table~\ref{tab4}). ArTS-\textit{ind} is the fine-tuned model to output a single trait name and score (e.g., Content 3). The results indicate that although the individual predictions highly outperform the baseline MTL model, our integrated method performs better on most traits. A single ArTS model outperforming 11 individual ArTS-\textit{ind} models is remarkable, highlighting our model's resource efficiency along with competitive performance. 

\paragraph{Feedback Prize dataset} 
To provide supplementary evaluation beyond traditional benchmarks and demonstrate generalizability across diverse datasets, we employ ArTS using the Feedback Prize dataset. The MTL model has not experimented with the dataset; accordingly, MTL* in Table~\ref{tab5} is our implementation results of the MTL with each trait scoring as the primary task and other traits as auxiliary tasks. Note that prompts are not differentiated, and all essays have identical traits in this dataset; therefore, the prompt number is excluded from the input, as in the ablation study. ArTS exhibits significantly improved QWK scores across all traits, demonstrating the broader applicability of ArTS (Table~\ref{tab5}). A greater improvement compared to the ASAP experiments further indicates that ArTS can yield a more substantial impact in the same trait composition settings compared to the multi-prompt and different trait scenarios. Furthermore, our single-model approach outperformed MTL* in predicting all six traits simultaneously, showcasing the efficiency of our model without the need for specialized auxiliary modules for each trait scoring.

\begin{figure}[t]
    \centering
    \includegraphics[width=7.5cm]{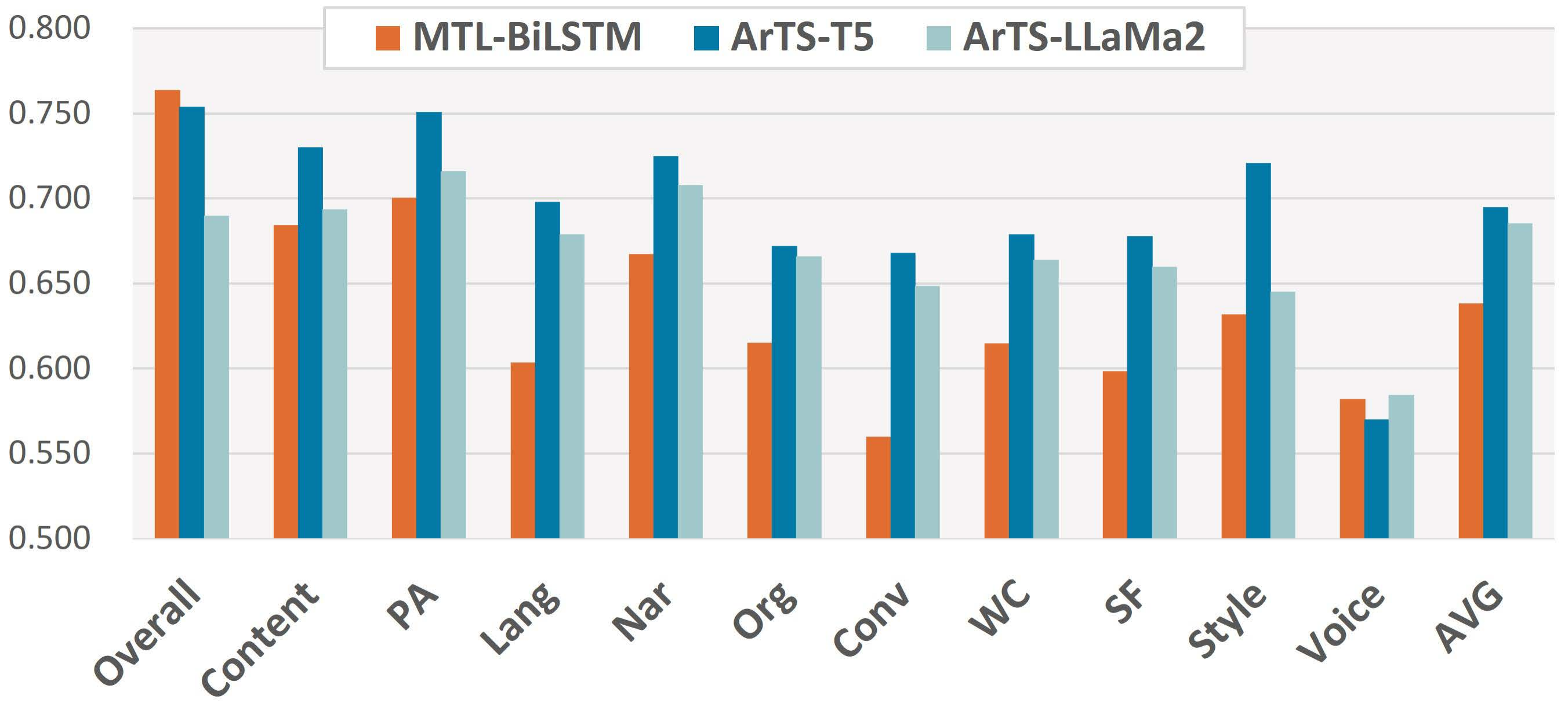}
    \caption{Results of ArTS with Llama2{\small-13B} and comparison with the baseline and ArTS with T5 models.}
    \label{fig3}
\end{figure}

\paragraph{Decoder-only LLM} 
To examine whether the decoder-only pre-trained language model alone could perform the function of autoregressive score generation, we fine-tuned the Llama2-13B model with our method (Figure~\ref{fig3}). Noticeably, ArTS-{\small Llama2} remarkably outperforms the baseline model for all the traits except for the \textit{Overall} score. However, ArTS-{\small T5} still performs better, suggesting the joint use of the encoder and decoder for AES. 

\paragraph{Comparison with BERT-based models}
% 최근 좋은 성능을 보이는 holistic scoring AES 연구들은 pre-trained BERT-based models을 사용해왔다. 우리가 아는 한, multi-trait scoring에 BERT-based model이 사용된 것이 없기에 overall score에 대해서만 성능을 비교해보았다. Overall score에 대한 그들의 성능은 0.79 to 0.805 정도로, 우리의 Overall score 채점 성능인 0.754보다 높은 것을 알 수 있다.

Recent studies in holistic AES have employed pre-trained BERT-based models and demonstrated promising scoring performances \cite{yang2020enhancing, cao2020domain, uto2020neural}. However, they have not been utilized in multi-trait scoring, which confines our performance comparison solely to the \textit{Overall} score. Their QWK results for the \textit{Overall} scoring range from 0.790 to 0.805 \citep{kumar2022many}, surpassing our 0.754. Our result aligns with the MTL model, exhibiting lower \textit{Overall} performance than BERT-based models but demonstrating training efficiency. Nevertheless, unlike MTL, we possess the advantage of simplicity and effectiveness by not requiring separate models for each prompt or trait and outperforming MTL in the other nine traits. %Furthermore, although ArTS exhibits a slightly lower \textit{Overall} score than MTL, its superior performance in other trait-scoring tasks (Table~\ref{tab2},~\ref{tab3}) suggests that our model may also excel over BERT-based models in those traits, particularly when data is limited.

% 우리 모델이 overall score에서 MTL보다 낮았지만, 다른 모든 trait에서 훨씬 성능이 좋았듯이, 우리 모델이 다른 trait scoring task, 특히 데이터가 더 적은 trait들에서는 BERT-based model 보다 좋을 수 있다. 

%Our results share the efficiency of MTL, which also showed lower Overall performance than BERT-based models, simultaneously predicting various trait scores with less time than BERT-based models.

Regarding training efficiency, using BERT-based models that predict a single numeric score for multi-trait predictions would require replicating multiple models, making it resource-inefficient. For example, predicting 11 traits with a BERT model of 110M parameters would involve a substantial 110M $\times$ 11 parameters, along with increased training time. This is a probable reason for the absence of a BERT-based system for multi-trait scoring tasks. In contrast, our approach enables multi-trait predictions across all prompts with a single T5-base model of 220M parameters, taking 16.3 hours for training time. When using T5-small of 60M parameters, which also highly outperforms the baseline model (Appendix~\ref{appen1}), it took about 2.8 hours for training. Unlike existing methods, which necessitate multiple trait-specific or prompt-specific models, the ArTS with a single model demonstrates both time and resource efficiency.

%하지만, 약 110 million 의 또한 MTL과 ~~에서 언급되었듯, BERT-base model은 그 자체로도 효율성 문제를 가진다. 

% 단 하나의 모델로 개별 모델 11개를 사용한 것보다 더 좋은 성능을 냈다는 것은 매우 놀랄 성적입니다. 

%\subsection{Fine-tuning for AES}
%We compare ArTS to the previous AES systems which utilize pre-trained models (Table~\ref{tab5}). As they are holistic scoring models, only the \textit{Overall} score is compared. It is noteworthy that our model of conducting multiple tasks outperforms holistic scoring models without combining additional strategies.

% pre-trained 모델을 사용함에 있어서 다른 많은 기법들을 적용시켰던 선행 연구들보다 우리의 간단한 접근법이 더 좋은 성능을 보였다는 점은 주목할만하다. 특히, 우리 모델이 다중 trait scoring을 동시에 처리하면서도 holistic scoring만 수행하는 모델을 뛰어넘었다는 것은 많은 것을 시사한다.       

\section{Conclusion}
In this paper, we introduce an autoregressive multi-trait scoring of essays that leverages the capacity of the pre-trained language model, T5. Our model exhibits remarkably improved results, demonstrating its ability to overcome far-lagging multi-trait-scoring performances. Furthermore, our approach allows a single model to make multi-trait score predictions across all prompts, avoiding the use of redundant modules and promoting simplicity and training efficiency. This indicates that a new paradigm of generating score sequences holds profound implications for future AES, opening new avenues for advanced multi-trait scoring.

\section*{Limitations}
We identified three limitations of this study. First, although our method achieved competitive results even in low-resource settings, it showed some performance degradation when confronted with extremely limited amounts of data, e.g., the \textit{Voice} trait with less than 1000 samples. This might be attributed to the inherent susceptibility of language models influenced by training data magnitude \cite{mehrafarin-etal-2022-importance}. Second, additional analysis regarding the prediction order can further enhance the scoring quality. Currently, the order is set from rare to frequent traits, which are decided by the number of rated prompts. In future work, we aim to explore more effective ordering strategies through detailed analysis. Lastly, a comprehensive exploration of other pre-trained models could shed more light on future AES. Previously, pre-trained models have only been applied for single-holistic scoring in AES. This could be attributed to the burdensome size of the pre-trained model to approach by constructing duplicated multiple trait-specific layers, unlike existing LSTM and attention-pooling-based models. Therefore, we could not directly compare our model to existing BERT-based systems for each trait scoring. However, as we have demonstrated the autoregressive approach to aid multi-trait AES, we plan to comprehensively investigate other alternative encoder-decoder or GPT-based models as the next step.

\section*{Acknowledgements}
This work was partly supported by Institute of Information \& communications Technology Planning \& Evaluation (IITP) grant funded by the Korea government (MSIT) (No.2022-0-00223, Development of digital therapeutics to improve communication ability of autism spectrum disorder patients, 50\%) and (No.2019-0-01906, Artificial Intelligence Graduate School Program (POSTECH), 50\%).

\bibliography{anthology,main}
\newpage
\appendix
% 각 trait 별로 성능 변화를 한 눈에 보기 위해 Table~\ref{tab2}에서의 main 실험 결과를 시각화해보았다. 전반적인 trait에 대해 성능이 올랐지만, Organization, Word Count, Sentence Fluency에서 특히 눈에 띄는 향상을 보임을 알 수 있다.  

\section{Effect of Model Size}
\label{appen1}

\begin{table*}[t]
\centering
\scalebox{
0.73}{
\begin{tabular}{l|ccccccccccc|c}
\hline
& \multicolumn{11}{c|}{\textbf{Traits (←)}} & \\
\hline
\textbf{Model} & Overall & Content & PA & Lang & Nar & Org & Conv & WC & SF & Style & Voice & AVG↑ (SD↓)\\
\hline
MTL{\small-BiLSTM} (baseline) & \textbf{0.764} & 0.685 & 0.701 & 0.604 & 0.668 & 0.615 & 0.560 & 0.615 & 0.598 & 0.632 & {0.582} & 0.638 (-) \\
\hline
ArTS-Small & 0.712 & 0.695 & 0.720 & 0.667 & 0.711 & 0.630 & 0.606 & 0.631 & 0.625 & 0.694 & 0.474 & 0.651 (±0.026)\\
ArTS-Base (Ours) & 0.754 & \textbf{0.730} & \textbf{0.751} & 0.698 & 0.725 & 0.672 & 0.668 & 0.679 & 0.678 & \textbf{0.721} & 0.570 & 0.695 (±0.018)\\
ArTS-Large & 0.751 & \textbf{0.730} & 0.750 & \textbf{0.701} & \textbf{0.728} & \textbf{0.675} & \textbf{0.682} & \textbf{0.680} & \textbf{0.680} & 0.715 & \textbf{0.603} & \textbf{0.700} (±0.024)\\
\hline
\end{tabular}}
\caption{\label{tab6}
Experimental results of fine-tuning ArTS with T5-Small, T5-Base, and T5-Large models. The left arrow (←) denotes the direction of trait prediction. Each value denotes the average QWK scores across all prompts for each \textbf{trait}.}

\smallskip
\smallskip

\scalebox{
0.75}{
\begin{tabular}{l|cccccccc|c}
\hline
& \multicolumn{8}{c|}{\textbf{Prompts}} & \\
\hline
\textbf{Model} & 1 & 2 & 3 & 4 & 5 & 6 & 7 & 8 & AVG↑ (SD↓) \\
\hline
MTL{\small-BiLSTM} (baseline) & 0.670 & 0.611 & 0.647 & 0.708 & 0.704 & 0.712 & 0.684 & 0.581 & 0.665 (-) \\
\hline
ArTS-Small & 0.696 & 0.669 & 0.682 & 0.732 & 0.712 & 0.743 & 0.712 & 0.492 & 0.680 (±0.029) \\
ArTS-Base (Ours) & \textbf{0.708} & \textbf{0.706} & 0.704 & \textbf{0.767} & 0.723 & \textbf{0.776} & \textbf{0.749} & 0.603 & 0.717 (±0.025)\\
ArTS-Large  & 0.701 & 0.698 & \textbf{0.705} & 0.766 & \textbf{0.725} & 0.773 & 0.743 & \textbf{0.635} & \textbf{0.718} (±0.030)\\
\hline
\end{tabular}}
\caption{\label{tab7}
Average QWK scores across all traits for each \textbf{prompt}.}
\end{table*}

We examine the impact of the pre-trained T5 model size (Table~\ref{tab6}). In additional experiments, we utilize T5-Small, T5-Base, and T5-Large, which contain 60 million, 220 million, and 770 million parameters, respectively. Experimental settings are all set as described in our main paper (Section~\ref{sec4}). 

For both trait-wise and prompt-wise results, overall performance improvements are observed as the model size increases. In particular, the \textit{Voice} trait with only 723 samples, including all training, development, and test sets, outperforms the baseline with ArTS-Large. This result highlights that utilizing larger models could boost the effect of our method, assisting even in severely low-resource environments.

% pre-trained T5 model의 사이즈에 따른 성능 변화를 살펴보았다. (따라 성능이 어떻게 달라지는지 살펴보았다.) 전반적으로 파라미터 크기에 따른 큰 변화는 없었지만 T5-Base와 Large가 small 모델보다 평균적으로 더 좋은 성능을 보였다. 구체적으로, 라벨링된 데이터가 많은 Overall, Prompt Adherence, Language, Organization, Conventions와 같은 trait일수록 Large 모델에서의 성능이 더 좋았고, Word Count, Sentence Fluency, Sytle, Voice 등 라벨링된 데이터가 더 적은 trait일수록 base 모델에서의 성능이 더 좋았다.    

\section{Comprehensive Results of Additional Experiments}
\label{appen2}

Due to the space constraint, only trait-wise results have been reported for additional experiments in Section~\ref{sec5}. In this section, we present both trait-wise and prompt-wise results for each experiment and numerical results for ArTS-Llama2, which are only shown in the graph figure.

\begin{table*}[t]
\centering
\scalebox{
0.73}{
\begin{tabular}{l|ccccccccccc|c}
\hline
& \multicolumn{11}{c|}{\textbf{Traits (←)}} & \\
\hline
\textbf{Model} & Overall & Content & PA & Lang & Nar & Org & Conv & WC & SF & Style & Voice & AVG↑ (SD↓)\\
\hline
MTL{\small-BiLSTM} (baseline)& \textbf{0.764} & 0.685 & 0.701 & 0.604 & 0.668 & 0.615 & 0.560 & 0.615 & 0.598 & 0.632 & {0.582} & 0.638 (-) \\
\hline
\textbf{ArTS} (Ours) & 0.754 & \textbf{0.730} & {0.751} & \textbf{0.698} & \textbf{0.725} & \textbf{0.672} & \textbf{0.668} & \textbf{0.679} & \textbf{0.678} & {0.721} & 0.570 & \textbf{0.695} (±0.018) \\
\hline
ArTS\textit{-rev} (→) & 0.739 & 0.724 & 0.749 & 0.687 & 0.718 & 0.667 & 0.658 & 0.660 & 0.666 & 0.711 & 0.562 & 0.686 (±0.021) \\
ArTS\textit{-ind} & 0.723 & 0.717 & \textbf{0.752} & 0.695 & 0.713 & 0.649 & 0.659 & 0.662 & 0.675 & \textbf{0.722} & 0.548 & 0.683 (±0.053) \\
\hline
ArTS\textit{-Llama2} & 0.690 & 0.694 & 0.716 & 0.679 & 0.708 & 0.666 & 0.649 & 0.664 & 0.660 & 0.645 & \textbf{0.584} & 0.685 (±0.034)\\
\hline
\end{tabular}}
\caption{\label{tab8}
Comprehensive results of models, which are described in Section~\ref{sec5}. Each value denotes the average QWK scores across all prompts for each \textbf{trait}. ArTS\textit{-rev} (→) predicts traits in reverse order, and 11 different ArTS\textit{-ind} models predict each trait individually. ArTS\textit{-Llama2} denotes the fine-tuned results of the Llama2-13B model. }

\smallskip
\smallskip

\scalebox{
0.75}{
\begin{tabular}{l|cccccccc|c}
\hline
& \multicolumn{8}{c|}{\textbf{Prompts}} & \\
\hline
\textbf{Model} & 1 & 2 & 3 & 4 & 5 & 6 & 7 & 8 & AVG↑ (SD↓) \\
\hline
MTL{\small-BiLSTM} (baseline) & 0.670 & 0.611 & 0.647 & 0.708 & 0.704 & 0.712 & 0.684 & 0.581 & 0.665 (-) \\
\hline
\textbf{ArTS} (Ours) & \textbf{0.708} & \textbf{0.706} & {0.704} & \textbf{0.767} & {0.723} & \textbf{0.776} & \textbf{0.749} & \textbf{0.603}& \textbf{0.717} (±0.025) \\
\hline
ArTS\textit{-rev} (→) & 0.700 & 0.683 & 0.702 & 0.763 & \textbf{0.730} & 0.767 & 0.734 & 0.586 & 0.708 (±0.027)\\
ArTS\textit{-ind} & 0.695 & 0.679 & \textbf{0.705} & 0.762 & 0.721 & 0.756 & 0.734 & 0.578 & 0.704 (±0.041)\\
\hline
ArTS\textit{-Llama2} & 0.702 & 0.641 & 0.700 & 0.721 & 0.691 & 0.736 & 0.700 & 0.592 & 0.685 (±0.030)\\
\hline
\end{tabular}}
\caption{\label{tab9}
Average QWK scores across all traits for each \textbf{prompt}.}
\end{table*}

\section{Error Analysis in Prompt Number Guidance}
\label{appen3}
In Section 5, we investigated the impact of providing a prompt when fine-tuning as an ablation study (Table~\ref{tab2},~\ref{tab3}). While the QWK results clearly demonstrated the effect of informing the prompt number, we conducted additional error case analysis. In particular, we find out that training with the \textit{"score the essay:"} prefix without providing a prompt (ArTS{\small-w/o \textit{Pr}}) often brings in \textit{out-of-range} scoring cases, influencing negatively on the overall QWK score. Each prompt has different score ranges for multiple traits, and we named the \textit{out-of-range} prediction for the prediction that is not inside the corresponding prompt's score range. While there are a five-fold total of 66 \textit{out-of-range} test predictions in ArTS{\small-w/o \textit{Pr}}, only one \textit{out-of-range} predictions are observed in ArTS. Note that ArTS is fine-tuned with the prefix \textit{"score the essay of the prompt N:"}. Most out-of-range cases are cases where an essay was mistaken for a different prompt and incorrectly graded based on the range of that prompt. The error case analysis proves that our strategy of prefixing with the prompt number provides clear evidence to the model about essay scoring, especially when there are numerous prompts.

\end{document}